# Image-Based and Sensor-Based Approaches to Arabic Sign Language Recognition

M. Mohandes, M. Deriche, and J. Liu

*Abstract*—Sign language continues to be the preferred method of communication among the deaf and the hearing-impaired. Advances in information technology have prompted the development of systems that can facilitate automatic translation between sign language and spoken language. More recently, systems translating between Arabic sign and spoken language have become popular. This paper reviews systems and methods for the automatic recognition of Arabic sign language. Additionally, this paper highlights the main challenges characterizing Arabic sign language as well as potential future research directions.

## I. INTRODUCTION

Sign language is the main form of communication among the deaf and the hearing-impaired. There are special rules of context and grammars that support the expression of a sign language. There are two main sign language recognition approaches: image-based and sensor-based. The main advantage of image-based systems is that signers do not need to use complex devices. However, substantial computations are required in the preprocessing stage. Instead of cameras, sensor-based systems use instrumented gloves equipped with sensors. However, sensor-based systems have their own challenges, including the cumbersome gloves worn by the signer.

To help people with normal hearing communicate effectively with the deaf and the hearing-impaired, numerous systems have been developed for translating diverse sign languages from around the world. Several review papers have been published that discuss such systems [1]–[9].

In [1], Wu and Huang presented a review of image-based gesture recognition. Different application systems, features, data collection methods, and recognition models were discussed. The authors showed that psycholinguistics, computer vision, and machine learning are all important in developing robust sign language recognition systems.

In [2], Pashaloudi and Margaritis presented a review of the hidden Markov model (HMM) for sign language recognition. The authors discussed the similarities and differences of using the HMM for speech and gesture recognition. Several glove-based and image-based systems were discussed.

In [3], Dipietro *et al.* provided a comprehensive survey of glove-based systems. Thirty types of gloves were discussed, outlining their characteristics and applications. These included the SayreGlove, MIT LED Glove, Digital Entry DataGlove, CyberGlove, and the PowerGlove, among others. The authors concluded that the DataGlove and

Manuscript received November 2, 2013; revised February 4, 2014 and April 11, 2014; accepted April 14, 2014. This work was supported by the King Fahd University of Petroleum and Minerals and the ASTF Grant ALJ-FFC07. This paper was recommended by Associate Editor G. Pirlo.
The authors are with the Department of Electrical Engineering, King Fahd University of Petroleum and Minerals, Dhahran 31261, Saudi Arabia (e-mail: mohandes@kfupm.edu.sa; mderiche@kfupm.edu.sa; liujunzhao@kfupm.edu.sa).
Color versions of one or more of the figures in this paper are available online at http://ieeexplore.ieee.org.
Digital Object Identifier 10.1109/THMS.2014.2318280

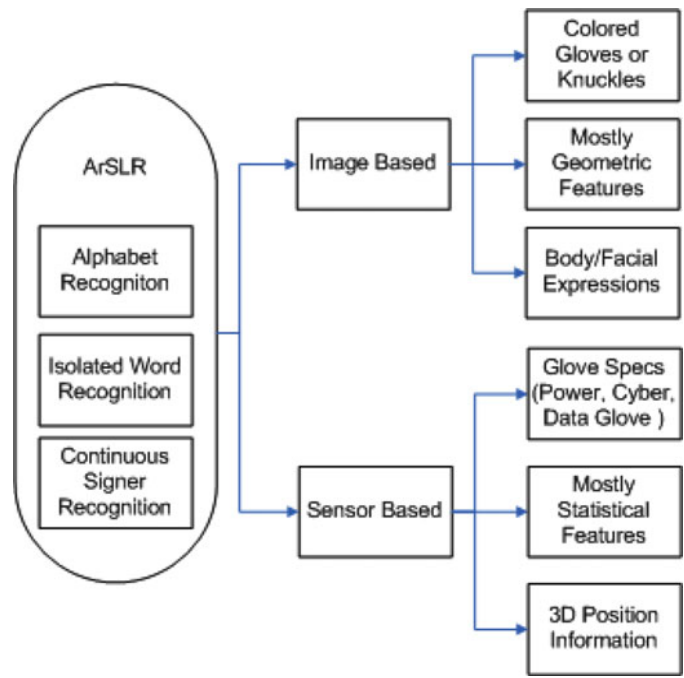

Fig. 1. Main classes of ArSLR algorithms.

the CyberGlove were the most commonly used gloves for sign language recognition.

In [4], Moni and Ali reviewed HMM-based techniques focusing mainly on systems using colored gloves for gesture recognition. They discussed the different approaches to decompose signs into sequences of hand gestures. The techniques reviewed in the paper focused on the detection of the hand region using edge detection algorithms to extract different geometric features.

In [5], Kausar and Javed discussed current trends in sign language recognition, categorizing existing algorithms into two classes: static and dynamic. They identified nine important issues in sign language recognition: segmentation, unrestricted environment, size of dictionary, invariance, variety of gestures, generality, start/end identification of gesture sequences, feature extraction, and feature selection. The paper concludes with a number of challenges and recommendations for future research in the field.

Most existing sign languages have been influenced by the French sign language system (FSL). However, communities from around the world had their own signing systems prior to the exposure to FSL [10]. The merging of these local signing systems with FSL led to unique sign languages for different communities. Each resulting sign language has now its own structure, grammar, syntax, semantics, pragmatics, morphology, and phonology. As such, systems should be developed to translate each sign language independently, including ArSL, as is currently done.

Compared with the work carried out for other sign languages [1]–[9], research on Arabic sign language recognition (ArSLR) has only witnessed a surge recently. Several methods and techniques have been used. Fig. 1 shows the most popular classes of ArSLR algorithms, highlighting the main features of the different approaches.

In this paper, a review of both image-based and sensor-based approaches is presented. In Section II, image-based ArSLR systems are presented followed by a discussion of sensor-based systems in Section III. In Section IV, future research directions and conclusions are detailed.

## II. Image-Based Arabic Sign Language Recognition

Traditionally, there have been three categories of image-based ArSLR systems: alphabet, isolated word, and continuous recognition. A typical image-based recognition system consists of five stages: image acquisition, preprocessing, segmentation, feature extraction, and classification. Earlier work focused on a limited vocabulary primarily used for basic human–machine interaction. The input to the image-based systems is a set of static images or a video sequence of signs. Usually, the signers are asked to briefly pause between signs for ease of separation. The main advantage of image-based ArSLR systems is user acceptance as the signer does not need to wear a cumbersome data glove. On the other hand, image-based techniques exhibit a number of challenges. These include: lighting conditions, image background, face and hands segmentation, and different types of noise, among others. Even though the segmentation of hands and face is computationally expensive, recent advances in computing and algorithms have made it possible to perform this segmentation in real time [11]. However, the widespread commercial deployment of image-based systems is still limited.

In the sequel, a discussion on the three main categories of recognition systems for Arabic sign languages is presented.

### A. Alphabet Recognition

Under this scenario, the signer performs each letter separately. Mostly, letters are represented by a static posture, and the vocabulary size is limited. In this section, several methods for image-based Arabic sign language alphabet recognition are discussed. The alphabet used for Arabic sign language is displayed in Fig. 2 [12].

Even though the Arabic alphabet only consists of 28 letters, Arabic sign language uses 39 signs. The 11 additional signs represent basic signs combining two letters. For example, the two letters "ال" are quite common in Arabic (similar to the article "the" in English). Therefore, most literature on ArSLR uses these basic 39 signs.

In [13], Mohandes introduced a method for automatic recognition of Arabic sign language letters. For feature extraction, Hu's moments were used followed by support vector machines (SVMs) for classification. A correct recognition rate of 87% was achieved.

Al-Jarrah and Halawani [14] developed a neuro-fuzzy system. The main steps of the system include: image acquisition, filtering, segmentation, and hand outline detection, followed by feature extraction. Bare hands were considered in the experiments, achieving a recognition accuracy of 93.6%.

In [15], Al-Rousan and Hussain built an adaptive neuro-fuzzy inference system for alphabet sign recognition. A colored glove was used to simplify segmentation, and geometric features were extracted from the hand region. The achieved recognition accuracy was 95.5%.

Assaleh and Al-Rousan [16] used a polynomial classifier to recognize alphabet signs. A glove with six different colors was used: five for fingertips and one for the wrist region. Different geometric measures such as lengths and angles were used as features. A recognition rate of 93.4% was achieved on a database of more than 200 samples representing 42 gestures.

In [17], Al-Jarrah and Al-Omari developed an image-based ArSL system that does not use visual markings. The images of bare hands are processed to extract a set of features that are translation, rotation, and scaling invariant. A recognition accuracy of 97.5% was achieved on a database of 30 Arabic alphabet signs.

In [18], Maraqa and Abu-Zaiter used recurrent neural networks for alphabet recognition. A database of 900 samples, covering 30 gestures performed by two signers, was used in their experiments. Colored gloves similar to the ones in [16] were used in their experiments. The

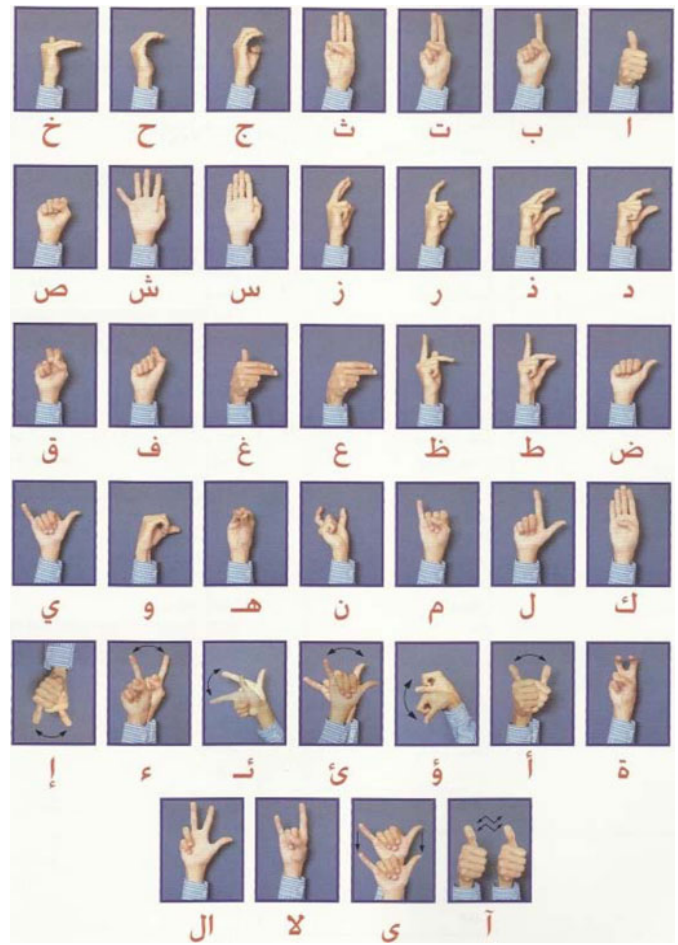

Fig. 2.　Arabic sign language alphabet.

Elman network achieved an accuracy rate of 89.7%, while a fully recurrent network improved the accuracy to 95.1%. The authors extended their work by considering the effect of different artificial neural network structures on the recognition accuracy. In particular, they extracted 30 features from colored gloves and achieved an overall accuracy of 95% [19].

In [20], El-Bendary et al. developed a sign language recognition system for the Arabic alphabet, achieving an accuracy of 91.3%. In their system, images of bare hands are processed. The input to the system is a set of features extracted from a video of signs, and the output is simple text. For each frame, the hand outline is first extracted. Using the centroid as a reference point, the distances to the outline of the hand covering 180° are extracted as a 50-D feature vector. These features are rotation, scale, and translation invariant. In the feature segmentation stage, they assumed a small pause between letters. Such pauses are used to separate the letter numbers and the related video frames. The signs of the alphabet are divided into three different categories before feature extraction. At the recognition stage, a multilayer perceptron neural network and a minimum distance classifier were used.

Hemayed and Hassanien [21] discussed an Arabic sign language alphabet recognition system that converts signs into voice. The technique is much closer to a real-life setup; however, recognition is not performed in real time. The system focuses on static and simple moving gestures. The inputs are color images of the gestures. To extract the skin blobs, the YCbCr space is used. The Prewitt edge detector is used to extract the hand shape. To convert the image area into feature vectors, principal component analysis (PCA) is used with a K-Nearest Neighbor Algorithm (KNN) in the classification stage.

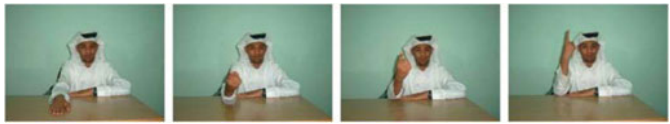

Fig. 3. Image sequence of the sign "1."

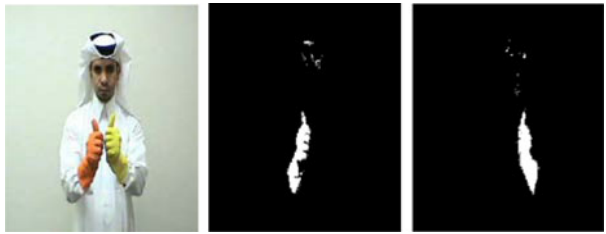

Fig. 4. Extracted right- and left-hand regions.

Naoum *et al.* [22] developed an image-based sign language alphabet recognition with an accuracy of 50% for bare hand, 75% for hand with a red glove, 65% for hand with a black glove, and 80% for hand with a white glove. The system starts by finding histograms of the images. Profiles extracted from such histograms are then used as input to a KNN classifier.

Elons *et al.* proposed a pulse-coupled neural network (PCNN) ArSLR system able to compensate for lighting nonhomogeneity and background brightness [23], [24]. The proposed system showed invariance under geometrical transforms, bright background, and lighting conditions, achieving a recognition accuracy of 90%.

In addition to the different image-based and glove-based systems that are currently in use, new systems for facilitating human–machine interaction have been introduced lately. In particular, the Microsoft Kinect and the leap motion controller (LMC) have attracted special attention. The Kinect system uses an infrared emitter and depth sensors, in addition to a high-resolution video camera. The LMC uses two infrared cameras and three LEDs to capture information within its interaction range. However, the LMC does not provide images of detected objects. The LMC has recently been used for Arabic alphabet sign recognition with promising results [25].

Arabic alphabet sign recognition is a relatively simple problem, the simplest among all image-based ArSLR approaches, as the vocabulary size is limited, and the signs are represented with mostly static images. Such systems achieve high recognition rates, typically over 90%. Note, however, that alphabet signs are not commonly used in daily practice. Their use is limited to finger spelling of words without specific signs like proper names. For these reasons, much of the current research efforts have been put into developing systems that focus on isolated words or even continuous sign recognition.

### B. Isolated Word Recognition

Unlike alphabet sign recognition, word sign recognition techniques analyze a sequence of images representing the entire sign, as shown in Fig. 3 [26].

In [27], Mohandes and Deriche used an HMM to identify isolated Arabic signs from images. They used a dataset consisting of 500 samples representing 50 signs. A Gaussian skin color model was used to find the signer's face that was then taken as a reference for hand movements. Two colored gloves (orange and yellow) were used for the right and left hands for ease of hand region segmentation (see Fig. 4). A simple region growing technique was used for hands segmentation. The recognition rate achieved over the 50 signs was 98%. In [28], the authors extended the work to cover a dataset of 300 signs, achieving a recognition accuracy of 95%.

Shanableh and Assaleh developed a signer-independent system for isolated Arabic signs [29]. They used segmented images of the hands extracted from colored gloves. For feature extraction, they used zonal discrete cosine transform (DCT) coefficients, while a KNN algorithm is used for classification. The authors achieved a classification rate of 87% over a vocabulary size of 23 signs. The same authors extended their work using HMM-based classification [30], [31]. They introduced new video-based features that take motion into account. The system achieved a recognition accuracy of about 95%.

In [32], Shanableh and Assaleh developed a video-based user-independent recognition system. The dataset consists of 3450 video segments covering 23 isolated gestures from three signers. The signers used colored gloves so that color information can be used in the preprocessing stage. Features are extracted from the accumulated differences of the images. A simple KNN algorithm is applied in the classification stage, achieving a recognition rate of 87%.

In [33], Youssif *et al.* developed an ArSLR system for isolated signs using the HMM. The regions of the palm and the fingers are modeled as ellipses and circles. They used a limited vocabulary size of 20 signs. With only eight features, they were able to achieve an accuracy of 82.2% in glove-free signer-independent mode.

Zaki and Shaheen [34] presented a combination of appearance-based features. Kurtosis position is used to identify the articulation location, while PCA is used to represent the hand region, and a motion code chain is used to represent the hand movement. With a database of 50 signs, the system achieved a recognition accuracy of about 90%.

In [35], Samir and Aboul-Ela proposed a semantic-oriented approach. Natural language processing rules were used to detect and correct errors from the classification stage. The proposed approach was shown to enhance recognition accuracy of ArSLR by around 20%.

In [36], Elons *et al.* used a PCNN for image feature generation from two different viewing angles. The features were evaluated using a fitness function to obtain a weighting factor for each camera. The features derived from the two images were used to obtain 3-D optimized features. The dataset used in the experiment contains 50 isolated words, and the achieved recognition accuracy was 96% for pose-invariant restrictions with a tolerance of up to 90°.

In [37], Al-Rousan *et al.* developed a system that was able to perform automatic translation of dynamic signs. The proposed hierarchical system divides signs into groups. For a given test sign, the group is first identified followed by recognition of the sign within that group. Twenty-three geometric features are used and tracked with an HMM classifier, achieving a recognition accuracy of 70.5% for user-independent mode and 92.5% for user-dependent mode. Their work was an extension of a previously developed algorithm that focused only on static postures [38].

In [39], Al Mashagba *et al.* developed an automatic isolated-word recognition system using two different-color gloves and an additional colored reference mark on the head. After extracting the three colored regions, five geometric features are extracted from any given video sequence. These features are: hand angle velocity, hand horizontal velocity, hand vertical velocity, hand horizontal position to the center of the head, and hand vertical position to the center of the head. A time delay neural network is used in the recognition stage, achieving a recognition accuracy of 77.4%.

Isolated word sign language recognition is more practical; however, it is more complex than alphabet recognition. More importantly, word recognition systems are required to deal with a sequence of images. The time component in analyzing such a sequence of images is very important. Note also that the vocabulary size for such systems can be very large. The challenge still remains in dealing with signs that are separated by certain pauses. It is observed that as the vocabulary size increases, accuracy decreases. For Arabic sign language, the size of the vocabulary needed for practical situations is still an open area for further research. In summary, the challenge for Arabic sign language is to develop signer-independent systems with a sufficiently large vocabulary size to make them suitable for practical deployment.

### C. Continuous Sign Language Recognition

While attractive in practice, continuous sign language recognition is more challenging than alphabet and isolated sign recognition. Such systems would be more representative of real-life signing situations for deaf people. The main challenge lies in detecting and modeling the extra movement resulting from the transition between the end of a certain sign and the start of the next one. Similar to speech recognition, most existing systems rely on the HMM for tracking time-varying patterns across the sequence of signs.

In [40], Assaleh *et al*. developed a user-dependent continuous ArSLR system for a database of 40 sentences composed of 80 commonly used words. Spatiotemporal features, based on the DCT, were used with an HMM classifier. The classifier was optimized with respect to the number of features, number of states, and number of Gaussian mixtures. A recognition accuracy of 94% was achieved with the optimized parameters.

Albelwi and Alginahi extended their initial alphabet recognition system to a real-time continuous ArSLR system [41]. After segmenting the hand regions, they used Fourier descriptors to model each hand's outer profile. For classification, they used a simple KNN algorithm and achieved a recognition accuracy of 90.6% on a limited size database.

In [42], Tolba *et al*. extended earlier works [43]–[45] by using PCNN and graph matching for continuous ArSLR. The experiments focused on sentences composed of three to four words. Signs are broken down into basic elements and static postures before applying the graph matching technique. The achieved recognition accuracy was above 70% for 30 continuous sentences composed of 100 gestures.

Research on continuous ArSLR is still limited compared with alphabet and isolated sign recognition. However, it has been observed that interest in such systems has been increasing. Continuous ArSLR systems are more relevant to practical situations for deaf people. An ideal continuous sign language recognition system should have a quick response in real time with a low error rate. Similar to speech processing, there is a need to develop systems that are based on phoneme-like subsigns as basic units. Such units can be used to enlarge the pool of signs as well as to enhance recognition accuracy.

### D. Other Related Work and Future Developments

Traditionally, research in sign language focused on systems that translate signs into speech or text. However, a more challenging and important problem is the translation of text or speech into signs. Such signs can be performed by avatars or generic 3-D models of the hands. Translation of text or speech into signs is important to complement sign language recognition for the full integration of the deaf into society. A number of attempts have been made to develop such systems, some of which focused on Arabic [46]–[49]. Furthermore, with the advances made in computing and mobile communications, some researchers have begun to develop sign language recognition systems for deployment over mobile platforms [50]–[52].

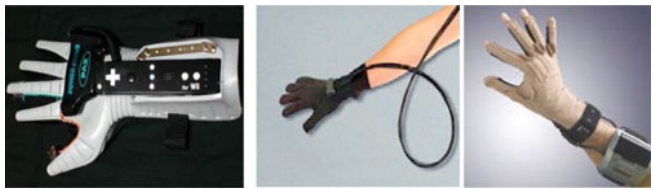

Fig. 5. Different Types of Gloves: PowerGlove (left), DT DataGlove (middle), CyberGlove (right).

### III. SENSOR-BASED ARABIC SIGN LANGUAGE RECOGNITION

Sensor-based recognition methods process data acquired from gloves equipped with sensors. The PowerGlove [53], DataGlove [54], [55], and CyberGlove [56] have commonly been used for ArSLR (see Fig. 5).

These gloves provide information on the position, rotation, movement, orientation of the hand, and more importantly, finger bending. A large number of features can be extracted from the data acquired from the gloves. These features can be used with a proper classifier to recognize the performed sign.

In [57] and [58], Mohandes *et al*. used a cost-effective off-the-shelf device to implement a robust ArSLR system. Statistical features were extracted from the acquired signals and used with an SVM classifier. With a database of 120 signs, a recognition accuracy of over 90% was achieved.

In [59], Assaleh *et al*. developed a low-complexity classification system. The glove used had five bend sensors and a 3-D accelerometer. From the acquired data, a number of statistical parameters were estimated. A regression technique was used to rank and select the most relevant features. The final list of selected features was used with a KNN classifier. With a database of ten signs performed by ten different signers, a recognition accuracy of 92.5% was achieved in the signer-independent mode, which increased to 95.3% for the signer-dependent scenario.

Ritchings *et al*. developed a computer-based system for teaching sign language using the DataGlove [60]. Bend sensors and push button switches were used to acquire 17 signals. The focus of the system was on assessing the ability of trainees to replicate signs performed by an expert signer. The database used covered 65 signs performed by four professional signers (teachers). The trainees were able to duplicate the signs with an accuracy of 93%.

In [61], a first attempt at two-handed Arabic sign recognition was made. The database consisted of 20 samples from each of 100 two-handed signs performed by two signers. Second-order statistics from subframes of the signs were used as features. The length of the feature vector was then reduced using PCA. For classification, an SVM classifier was used, achieving an accuracy of 99.6% with 100 signs.

In [62], Mohandes and Deriche used the Dempster-Shafer Theory of Evidence to combine decisions from the CyberGlove and a hand tracking system. The authors showed that fusion at the decision level outperforms traditional feature-based fusion. They started with some basic experiments using the CyberGlove and the hand tracking systems separately. The hand tracking system achieved an accuracy of 84.7%, while the CyberGlove system achieved an accuracy of 91.3%. The traditional feature-based combination of the two systems provided a maximum accuracy of 96.2%. This accuracy was then improved to 98.1% when fusion was carried out at the decision level.

### IV. DISCUSSION AND CONCLUSION

Although research in sign language recognition began several decades ago, it is still in its infancy, as no such systems have been deployed on a large scale to date. Research in this area will undoubtedly impact other applications involving human–machine interaction.

As outlined in this review, two major approaches have been used in translating sign language: sensor-based and image-based techniques. This paper summarized the state of the art for both approaches with particular focus on Arabic sign language. As was shown, both approaches have their own advantages and disadvantages. The ultimate goal of systems translating between deaf and vocal people is to facilitate communication in a restriction-free environment without requiring the signer to wear cumbersome devices or colored gloves. Researchers continue to put substantial efforts into developing systems that ease these restrictions. Microsoft Kinect, for example, has recently been used as an interface for sign language recognition [63], [64]. However, it has not been used widely for ArSLR.

We have seen that current research in ArSLR has only been satisfactory for alphabet recognition, with accuracy exceeding 98%. Isolated Arabic word recognition has only been successful with medium-size vocabularies (less than 300 signs). On the other hand, continuous ArSLR is still in its early stages, with very restrictive conditions. Our survey showed that major efforts are needed to reach the aforementioned objective. In what follows, we discuss the major challenges facing the development of ArSLR systems for mass deployment.

If we consider the important criterion of reducing the restrictions on the environment, sensor-based systems would be the obvious choice. For sensor-based systems, research efforts have focused on two main directions: selection of the appropriate glove for a given application and the development of robust signal processing tools.

Three main issues need to be considered for selecting the appropriate glove for the given application. The first one is related to the number and location of sensors, which directly impacts the size of the dictionary. A second problem is related to the lack of detailed analysis of the different system's components, including the sensors, the support, and the electronics. The last issue is related to calibration, as different people have different hand sizes and finger length/thickness. As a consequence, glove sensors may not be aligned with finger joint locations. To reduce inaccuracies, gloves need to be calibrated for a particular user. This raises the issue of whether sensor-based systems are more suitable for signer-dependent scenarios.

The second direction of major importance is the signal processing stage of SLR systems. Traditionally, SLR systems have been seen as typical pattern recognition systems. Such systems involve preprocessing, feature extraction, and classification. The extraction of features from noisy data can be a major challenge especially when we consider that the patterns are represented with nonstationary signals (dynamic gestures). The classification stage can also be a major challenge when we consider vocabularies consisting of both single-hand and two-handed signs. HMMs, in particular, are found to be inappropriate for coarticulation. Fusing information from the different sensors can also be tackled at different levels. These include the data, feature, and decision levels.

For practical deployment of sign language translation systems, studies have shown that image-based systems are more attractive to users than sensor-based systems. While sensor-based methods require signers to wear data gloves connected to specialized signal-processing boards, image-based methods do not have this limitation. Additionally, image-based ArSLR systems can benefit from complementary information obtained from facial expression and head/lip movements. However, this additional information is not currently used widely for ArSLR systems. Moreover, image-based approaches still require special setup for acquiring the signs. These include: background, lighting, signer cloths, and camera(s). All of these factors have a major impact on the overall performance of image-based systems.

With the advances made in computing, it is now possible to combine information from traditional approaches with the aforementioned complementary information in real time. Research in this field can further be expanded to cover grammatical interpretation of sequences of signs. This will prove to be very useful in developing natural signing systems. Such systems should convey both temporal and spatial information. Spatial information in particular is very important in natural setups as signs can point to locations previously established as reference positions.

With the use of more than one camera, 3-D data can enhance the detection capabilities of the system as well as the size of the vocabulary. Multicamera systems may help in lessening some of the environmental restrictions. However, 3-D models bring additional computational load which can be handled by the rapid advancement of computing systems.

To enhance the performance of sensor-based and image-based systems, some researchers have started looking into hybrid approaches that combine information from cameras as well as gloves. The availability of powerful computing systems makes such hybrid systems feasible. A number of issues can be investigated by using this combination. These include synchronization and the type of information for fusion. In particular, fusion can be performed at several levels, including the data, feature, and decision levels. In addition to the above, linguistic grammars can further enhance the performance of such systems.

As smart mobile devices become widely accessible, we can expect to see more sign language translation systems deployed on such platforms. While it is important to further enhance the performance of existing sign to speech translation systems, a bigger challenge is to develop systems able to translate text or voice into signs performed by avatars or human signers on the same portable devices. Developing such systems will also contribute to two-way communication across sign languages. The intensive research efforts in sign language recognition paired with advances in technology are expected to facilitate the full integration of the deaf community into the rest of society. The scope for improving existing systems is still wide open especially for Arabic sign language.

Finally, while excellent results have been achieved in ArSL recognition under simple scenarios and environmental setup, a major challenge still resides in developing systems that provide robust performance under minimum restrictions. To face this challenge, there is a need to consider hybrid systems that combine not only multiple algorithms, but also nonhomogeneous sensors like cameras, sensors, LMC, Kinect, and so on. Such systems are expected to translate ArSL in real time with the least restriction and with high accuracy.